# Semi-Supervised Text Categorization using Recursive K-means clustering

Harsha S Gowda, Mahamad Suhil, D S Guru and Lavanya Narayana Raju

Department of Studies in Computer Science, University of Mysore, Mysore, India.
`harshasgmysore@gmail.com, mahamad45@yahoo.co.in, dsg@compsci.uni-mysore.ac.in and swaralavz@gmail.com`

**Abstract.** In this paper, we present a semi-supervised learning algorithm for classification of text documents. A method of labeling unlabeled text documents is presented. The presented method is based on the principle of divide and conquer strategy. It uses recursive *K*-means algorithm for partitioning both labeled and unlabeled data collection. The *K*-means algorithm is applied recursively on each partition till a desired level partition is achieved such that each partition contains labeled documents of a single class. Once the desired clusters are obtained, the respective cluster centroids are considered as representatives of the clusters and the nearest neighbor rule is used for classifying an unknown text document. Series of experiments have been conducted to bring out the superiority of the proposed model over other recent state of the art models on 20Newsgroups dataset.

## 1 Introduction

The amount of text content available over the web is so abundant that anybody can get information related to any topic. But, the performance of retrieval systems is still far below the level of expectation. One of the major reasons for this is the amount of labeled text available is negligibly small when compared to that of the unlabeled text. Thus, automatic text categorization has received a very high demand by many applications to well organize a huge collection of text content in hand. To this end, many machine learning based algorithms are being developed to best make use of the unlabeled text in addition to the available labeled text to draw clear boundaries between different classes of documents present in the corpus (Nigam et al., 2011; aZhang et al., 2015; Su et al., 2011). Hence, further categorization of unlabeled samples can be done effectively. The process of using unlabeled samples along with a small set of labeled samples to better understand the class structure is known as semi-supervised clustering (Zhu., 2008).

Generally there are two approaches to semi-supervised learning; one is a similarity based approach and the other one is a search based approach. In similarity based approach an existing clustering algorithm that uses a similarity matric is employed, while in search based approach, the clustering algorithm itself is modified so that the user provided labels are used to bias the search for an appropriate partition. A detailed comparative analysis on these two approaches can be found in (Basu et al., 2003). Further, readers can find a review on semi supervised clustering methods for more details (Bair, 2013).

A method based on hierarchical clustering approach is proposed by (Zhang et al., 2015) where labeled and unlabeled texts are respectively used for capturing silhouettes and adapting centroids of text clusters. A simple semi-supervised extension of multinomial Naive Bayes has been proposed (Su et al., 2011). This model improves the results when unlabeled data are added. The notion of weakly related unlabeled data is introduced in (Yang et al., 2009). The strength of this model is that it works even on a small training pool. A model based on combination of expectation maximization and a Naive Bayes classifier is introduced by (Nigam et al., 2013). This algorithm trains a classifier using available label documents first and subsequently labels the unlabeled documents probabilistically. A variant of expectation maximization by integrating Bayesian regression is also proposed (Zhang et al., 2015). Based on co-clustering concept a fuzzy semi supervised model can also be traced in literature (Yan et al., 2013).

Based on the above literature survey we learnt that the notion of semi-supervised learning is receiving greater attention by the researchers in recent years. It is also noted that consideration of unlabeled data at the time of learning along with labeled data has a tendency of improving the results. In this direction, here in this work, we made a successful attempt for text categorization through semi-supervised learning and as a result of it, we propose a search based approach. The proposed approach is based on a partitional clustering where in, the *K*-means algorithm is tuned up to be a recursive algorithm. The proposed model works based on divide and conquer strategy. Initially *K*-means clustering algorithm is used to partition the sample space into as many as the number of classes and subsequently, on each obtained cluster we employ *K*-means algorithm recursively till the partitions meet a pre-defined criterion.

Rest of the paper is organized as follows: Section 2 presents the detailed description of the proposed model for labeling of unlabeled samples through recursive semi-supervised *K*-means clustering. In section 3, the representation of documents, experimental setup including datasets and evaluation measures have been presented. Section 4 provides the results and analysis of the proposed model on different datasets. Finally section 5 presents the conclusion of the paper followed by references.

## 2  Proposed Method

The proposed model has two major stages; learning, through recursive *K*-means clustering and classification, to label a given unknown text document.

## 2.1 Recursive K-means Clustering

Let $D = \{D^L, D^U\}$ be a collection of $N$ text documents where, $D^L = \{d_1^l, d_2^l, d_3^l, ..., d_{N_1}^l\}$ be the set of $N_1$ labeled documents and $D^U = \{d_1^u, d_2^u, d_3^u, ..., d_{N_2}^u\}$ be the set of $N_2$ unlabeled documents (where, $N_1 << N_2$ and $N_1 + N_2 = N$). Let $C = \{C_1, C_2, ...., C_K\}$ be the set of $K$ classes present in $D^L$. In semi-supervised clustering, the task is to label the documents in $D^U$ with one of the $K$ different class labels using documents of $D^L$. For this purpose, we initially consider the labeled documents along with a subset of $D^U$ with $N_3$ number of randomly chosen documents say $D^L$ to form a training collection $D^T = \{D^L, \bar{D}^U\}$ with $N_{tr}$ ($N_1+N_3$) documents. In the next subsection, a recursive $K$-means clustering algorithm is proposed to cluster the documents in $D^T$ into many partitions consisting of both unlabeled and labeled documents such that each partition should contain labeled samples from a single class.

Initially $K$-means clustering is applied on $D^T$ with $K=K$ (the number of classes) since $D^T$ has labeled samples of all $K$ classes. If a partition $P_i$ has labeled samples from more than one class then $K$-means is applied again with $K$ being the number of distinct classes present in $P_i$. This process is applied recursively on each sub partition till the entire training collection $D^T$ is partitioned into many small clusters say $M$ in number and represented by $FinalClusters = \{P_1, P_2, ..., P_M\}$ such that each cluster contains labeled samples from strictly a single class. So, this recursive $K$-means is a based on the divide and conquer strategy. The major difference is that, divide and conquer normally partitions the problem into predefined number sub-problems at each stage, whereas, the proposed recursive $K$-means decides the number of sub-problems dynamically based on the number of unique classes present in each cluster. Hence, we may end up with multiple clusters for each class depending on the variations present within the class. Let $ClusterLabels = \{l_1, l_2, ..., l_M\}$, where $l_m \in [C_1, C_K]$, be the class labels of each cluster in the FinalClusters set. Fig 1 shows a simple illustration of the recursive $K$-means clustering on a 4 class problem.

Further, every unlabeled sample of $D^T$ is labeled by the class label of its respective cluster as their final labels. The centroids of respective clusters say $ClusterCentroids = \{PC_1, PC_2, ..., PC_M\}$ are computed as mean of each cluster. A knowledgebase of *FinalClusters*, *ClusterCentroids* and *ClusterLabels* is then created for the purpose of labeling of unlabeled samples in $D^U$ during classification. The proposed recursive $K$-means clustering for semi-supervised learning is detailed in Algorithm 1.

Input to the recursive $K$-means algorithm is a collection of labeled and unlabeled samples, $D^T$. In Initialization part, the number of unique labels present in $D^T$ is set to $K$, and $K$-means clustering is performed by randomly choosing $K$ samples, each from a different class, as initial seeds to arrive at $K$ disjoint clusters of the data say

*Partition*. The process of recursive *K*-means clustering is applied to each cluster in *Partition* as explained below.

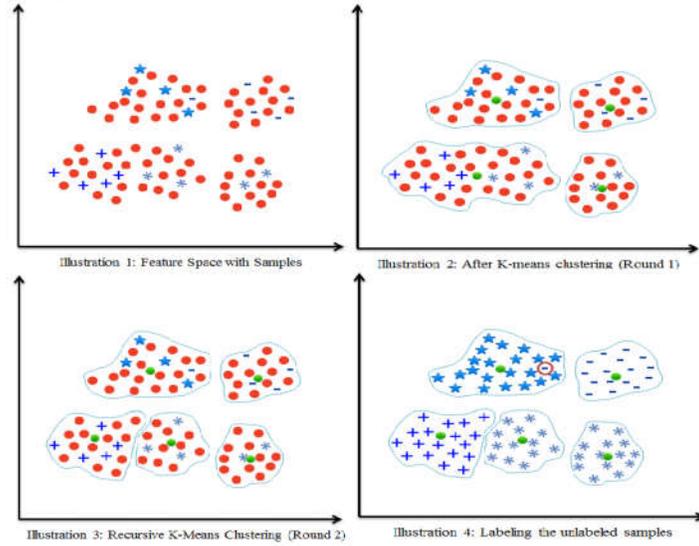

**Fig 1.** Illustration of the proposed recursive *K*-means clustering for semi-supervised learning.

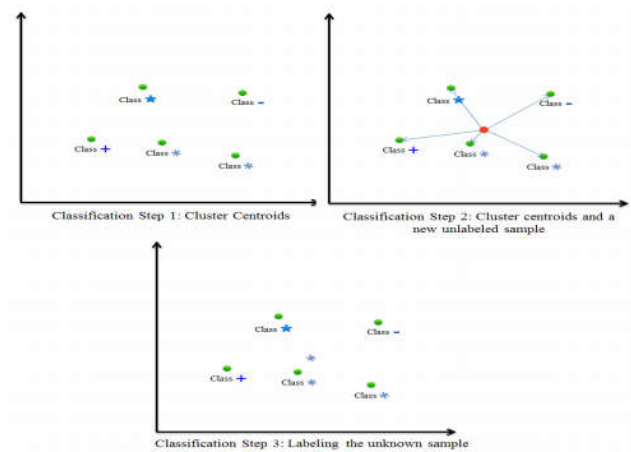

**Fig 2.** Labeling of an unlabeled sample

For every cluster $P_i$ in *Partition*, the unique class labels are computed. The label of a class with maximum labeled samples out of all the labeled samples in $P_i$ is assigned as the *ClassLabel* of $P_i$. If $P_i$ has labeled samples from two or more classes and if the ratio of the number of samples from a class other than class of $P_i$ to the number of samples from the class of $P_i$ is greater than a predefined threshold '*Th*'; then $P_i$ is treated as a new collection of labeled and unlabeled samples and hence the Algorithm1 is re-invoked with $K$ being the number of unique class labels in $P_i$. The threshold *Th* is included to eliminate the outliers, which is set empirically. The clusters which do not undergo this recursive clustering are added to the *FinalClusters*.

Once, all the clusters in Partition are processed by the proposed recursive K-means clustering, all the unlabeled sample of each cluster in FinalClusters are assigned by the ClassLabel of the respective cluster.

---

**Algorithm 1: Recursive *K*-Means for Semi-Supervised Learning: RKMSSL**

Input:
$D^T = \{D^L, \overline{D}^U\}$ : A collection of labeled and unlabeled samples.

Output:
*FinalClusters* : a collection of labeled clusters
Labels for the unlabeled samples in $D^T$

Method
  **(i). Initialization**
Step 1.    $K \leftarrow$ Number of unique labels in $D^T$ (number of classes)
Step 2.    *InitialSeeds* $\leftarrow$ randomly chosen $K$ samples (one from each unique class present in $D^T$)
Step 3.    *SeedLabels* $\leftarrow$ cluster indices of *IntialSeeds*

  **(ii). Initial *K*-means clustering**
Step 4.    Apply *K*-means on $D^T$ with *IntialSeeds* as initial cluster centroids to get K clusters say, *Partition*

  **(iii). Recursive K-means clustering**
Step 5.    For every cluster $P_i$ in *Partition*
Step 6.      $NCP_i \leftarrow$ Number of unique class labels in $P_i$ and
      $LSP_i \leftarrow$ Number of labeled samples of each unique class in $P_i$
Step 7.      If $NCP_i > 1$ then,
Step 8.       $ClassLabel(P_i) \leftarrow$ Label (max ($LSP_i$))
Step 9.       For each class $j$ other than in $ClassLabel(P_i)$
Step 10.       $RelativePercentage_j \leftarrow (LSP_i [j] / LSP_i [ClassLabel(P_i)])*100$
Step 11.       If $RelativePercentage_j >$ greater than *Th* then,
Step 12.        RKMSSL ($P_i$)
Step 13.        Goto step 15;
      end
     End for
     *Add $P_i$ to the FinalClusters* set
Step 14.    End if

  **(iv) Labeling**
Step 15.    For every unlabeled document $d_u$ in cluster $P_i$
Step 16.      $Label(d_u) \leftarrow ClassLabel(P_i)$
Step 17.    End for
Step 18. End for
Algorithm Ends

```
Algorithm 2: Classification
  Input:
            FinalClusters : a collection of labeled clusters
            ClusterLabels: class labels of each cluster in FinalCluster
            D^U: Collection of unlabeled samples
  Output:
            ComputedLabels: Labels for the unlabeled samples in D^U
  Method:
  Step 1.   for every document d_u in D^U
  Step 2.      for every cluster F_c in FinalClusters
  Step 3.         Dist(d_u,F_c)=distance(d_u,Centroid(F_c))
  Step 4.      End for
  Step 5.      ComputedLabel(d_u)=ClusterLabels(argmin(Dist))
  Step 6.   End for
```

## 2.2 Classification

Given a collection of unlabeled samples $D^U$, they are classified into one of the *K* classes by comparing with the centroids of the clusters formed in the learning stage using Algorithm 2. Initially, the distance of an unlabeled sample du in $D^U$ to the centroids of all the clusters in *FinalClusters* is computed. Then $d_u$ is labeled by the label of a cluster which has the least distance to it or simply the nearest neighbor classifier is employed on the cluster centroids to label an unknown document. An illustration of labeling a given unlabeled sample is shown in Fig 4.Representation of Documents in Lower Dimensional Space.

## 3  Experimental Setup

### 3.1  Representation of Documents in Lower Dimensional Space

In this section, we present the representation scheme followed for the text documents. The importance of this section is that, it has to be brought into the notice of the reader that we do not represent the documents using conventional vector space model (VSM) using the bag of words (BoW) constructed for the entire collection. This is due to the fact that, VSM leads to a very high dimensional sparse matrix which is not effective if directly used in computations and hence dimensionality reduction has to be applied (Sebastiani., 2003). To alleviate this problem, (Isa et al., 2008) have proposed an effective text representation scheme which can reduce the dimension of the documents equal to the number of classes at the time of representation itself. Recently, we can track a couple of attempts in the literature which have proven the effectiveness of this representation in addition to its time efficiency (Guru et al., 2010, Harish et al., 2010). Besides, (Guru and Suhil., 2015) have proposed a novel term-weighting scheme called 'Term_class Relevance Measure (TCR)' to be used with the representation scheme of Isa et al., (2008) for achieving better performance. Hence, we adapt the representation from Isa et al., (2008) with term weighting scheme of (Guru and Suhil., 2015).

### 3.2 Dataset and Evaluation measures

To validate the applicability and efficiency of the proposed method, we conducted a series of experiments on 20Newsgroups dataset. Since we are using supervised datasets to evaluate the performance of the proposed method, we use the well-known information retrieval measures to quantize the results obtained viz., Precision (P), Recall (R) and F-measures (F) in both micro and macro averaging (Manning et al., 2008). The original 20Newsgroups collection is nearly balanced corpus consists of 18846 documents from 20 different news topics. In our experiments we have considered the documents from top 10 categories of the 20Newsgroups to form a dataset of 9645 text documents. We divided the entire dataset into two halves where the first half was used for model building and the second half is used for classification which we call respectively as 'training set' and 'test set'. The first half is again divided into two subsets; one is a very small subset of labeled samples and the other one is a large subset of unlabeled. We varied the ratio of labeled and unlabeled from 1:49 to 20:30 in steps of 1 and hence 20 different cases were experimented. For each case, 20 different random trials were considered to study the consistency of the results obtained by the proposed method on the test set.

## 4 Results and Analysis

In this section, we present the results and analysis of the various experiments conducted on 20Newgroups dataset to evaluate the proposed method.

Fig 3 shows the performance of the proposed method for varying percentage of labeled samples in terms of classification accuracy. It can be observed from the Fig 3 that the average accuracy of the proposed method is very high and consistent under varying percentage of labeled samples though it has attained a minimum value for some trails. Fig 4 and Fig 5 show respectively the precision and recall curves of the proposed method for varying percentage of labeled samples in terms of maximum, minimum, average and standard deviations of the 20 random trails conducted. The high average values and the less standard deviation among the values of different trials indicate that the proposed method is consistent in addition to being very efficient except a few cases where the performance has declined to a minimum value. Furthermore, performance in terms of macro and micro averaged F-measure values has also been studied and are respectively shown in Fig 6 and Fig 7. The similar observation as in precision and recall can also be drawn for macro and micro averaged F-measures. Moreover the similarity in the values of macro and micro averaged F-measures validate the fact that the 20Newsgroups dataset is a balanced corpus.

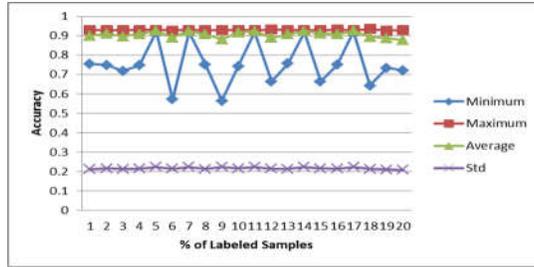

**Fig 3.** Accuracy of the proposed method for varying percentage of labeled samples

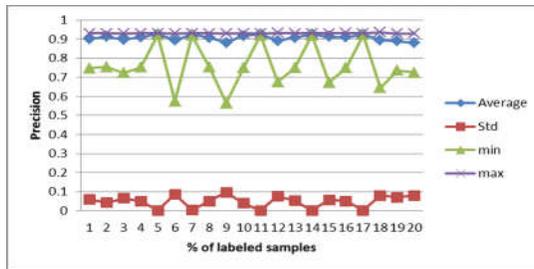

**Fig 4.** Macro averaged precision of the proposed method for varying percentage of labeled samples

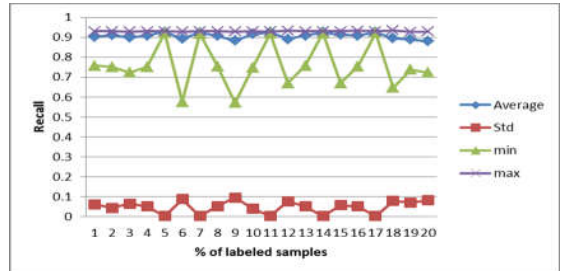

**Fig 5.** Macro averaged recall of the proposed method for varying percentage of labeled samples

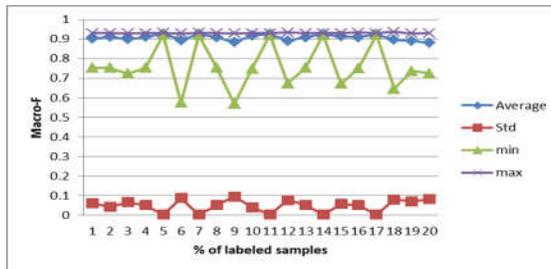

**Fig 6**. Macro averaged F-measure of the proposed method for varying percentage of labeled samples

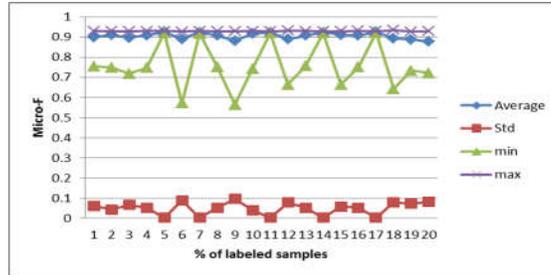

**Fig 7**. Micro averaged F-measure of the proposed method for varying percentage of labeled samples

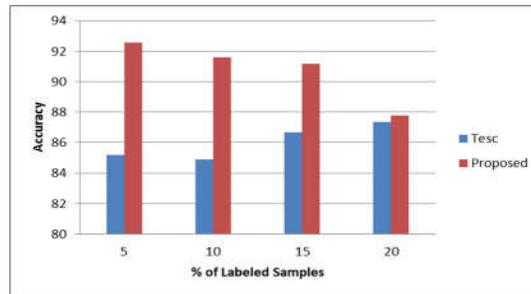

**Fig 8**. Comparison of the proposed method with the TESC (Zhang et al., 2015) for varying percentage of labeled samples in terms of Accuracy.

Besides, a quantitative comparative analysis of the proposed method is also made with one of the recently proposed semi-supervised method TESC (Zhang et al., 2015) which has shown very good performance for text categorization applications. The other reason for the consideration of TESC for comparison is that TESC follows the similar basic principle of arriving at clusters by using unlabeled samples along with a small set of labeled samples. Further the unlabeled samples are labeled in a similar fashion followed by the proposed method, i.e., a test sample is labeled as a member of a class by comparing it with the centroids of each cluster formed during training process. The major difference lies in the proposed method when compared to TESC is that, they follow an approach very similar to hierarchical clustering to arrive at the clusters whereas, the proposed method proposes a recursive *K*-means clustering. We conducted the experiments on TESC with four different percentages of labeled samples viz., 5%, 10%, 15% and 20% respectively and are compared with that of the proposed method using accuracy as shown in Fig 8. It can be clearly observed from the Fig 8 that, the proposed method outperforms TESC in all the four cases. Another observation is that the proposed method can learn effectively with a small quantity of labeled samples itself whereas TESC requires sufficiently large number of labeled samples to learn the silhouettes of the clusters effectively.

## 5 Conclusions

A semi-supervised model for classification of text documents is presented. The model works on divide and conquer strategy and uses K-means algorithm recursively. The finding of the work is on adaptation of partitional clustering algorithm for semi-supervised learning of text documents.

**Acknowledgements**

The second author of this paper acknowledges the financial support rendered by the University of Mysore under UPE grants for the High Performance Computing laboratory. The first and fourth authors of this paper acknowledge the financial support rendered by Pillar4 Company, Bangalore.